\documentclass[lettersize,journal]{IEEEtran}
\usepackage{amsmath,amsfonts}
\usepackage{algorithmic}
\usepackage{algorithm}
\usepackage{array}
\usepackage[caption=false,font=normalsize,labelfont=sf,textfont=sf]{subfig}
\usepackage{textcomp}
\usepackage{stfloats}
\usepackage{url}
\usepackage{verbatim}
\usepackage{graphicx}
\usepackage{cite}
\usepackage{threeparttable}
\usepackage[utf8]{inputenc} 
\usepackage[T1]{fontenc}    
\usepackage{hyperref}       
\usepackage{url}            
\usepackage{booktabs}       
\usepackage{amsfonts}       
\usepackage{nicefrac}       
\usepackage{microtype}      
\usepackage{xcolor}         
\usepackage{colortbl}   
\usepackage{makecell}
\usepackage{graphicx}
\usepackage{amsmath}
\usepackage{multirow}
\usepackage{xspace}
\usepackage{wrapfig}
\usepackage{pifont}
\usepackage{wrapfig}
\usepackage{tcolorbox}
\tcbuselibrary{skins}
\usepackage{listings}
\definecolor{indigo}{HTML}{4B0082}
\definecolor{ourlightblue}{HTML}{C7EBFF}         
\definecolor{ourlightorange}{HTML}{FFDBC3}       
\definecolor{ourlightgray}{HTML}{EAEAF2}         
\definecolor{ourlightgreen}{HTML}{BDF0E2}        
\definecolor{ourdarkblue}{HTML}{0033CC}          
\definecolor{softblue}{RGB}{230,234,250}
\definecolor{boldblue}{RGB}{196,210,248}
\definecolor{chocolate}{HTML}{D2691E}

\newcommand{\bc}[1]{\cellcolor{boldblue}#1}

\newcommand{\AdaSteer}{{\protect\color{chocolate} ${{\textbf{\textsc{\texttt{CoDA}}}}}$}\xspace}
\newcommand{\avgbg}{\cellcolor{gray!15}}
\newtcolorbox[auto counter, number within=section]{promptbox}[2][]{
  colback=white,
  colframe=blue!80!black,
  coltitle=white,
  title=Prompt~\thetcbcounter: #2,
  fonttitle=\bfseries\normalsize,
  boxrule=1pt,
  arc=2mm,
  top=2mm,
  bottom=2mm,
  width=0.95\textwidth,
  #1
}

\newtcolorbox[auto counter, number within=section]{examplebox}[2][]{
  colback=white,
  colframe=black,
  boxrule=0.5pt,
  sharp corners,
  left=4pt,
  right=4pt,
  top=4pt,
  bottom=4pt,
  enhanced,
  title=Example~\thetcbcounter: #2,
  #1
}

\def \VersionWithComments {}
\ifdefined 
\VersionWithComments
\usepackage{marginnote}

\makeatletter
\DeclareRobustCommand\onedot{\futurelet\@let@token\@onedot}
\def\@onedot{\ifx\@let@token.\else.\null\fi\xspace}

\begin{document}

\title{CoDA: Towards Effective Cross-domain Knowledge Transfer via \underline{Co}T-guided \underline{D}omain \underline{A}daptation}

\author{Jianzhi Yan, Le Liu, Buzhou Tang, Yang Xiang, Dongning Sun, Zhiming Li
    
\thanks{Jianzhi Yan is with Harbin Institute of Technology, Shenzhen, 518100, China, and also with Pengcheng Laboratory, Shenzhen, 518100, China (e-mail:yanjzh@pcl.ac.cn)
}
\thanks{Le Liu is with Harbin Institute of Technology, Shenzhen, 518100, China, and also with Pengcheng Laboratory, Shenzhen, 518100, China (e-mail:liul07@pcl.ac.cn)
}
\thanks{Buzhou Tang is with Harbin Institute of Technology, Shenzhen, 518100, China, and also with Pengcheng Laboratory, Shenzhen, 518100, China (e-mail:tangbuzhou@gmail.com)
}
\thanks{Yang Xiang is with Pengcheng Laboratory, Shenzhen, 518100, China (e-mail:xiangy@pcl.ac.cn)
}
\thanks{Dongning Sun is with Pengcheng Laboratory, Shenzhen, 518100, China (e-mail:sundn@pcl.ac.cn)
}
\thanks{Zhiming Li is with Pengcheng Laboratory, Shenzhen, 518100, China (e-mail:lizhm02@pcl.ac.cn)
}
\thanks{\textit{(Equal contribution: Jianzhi Yan; Zhiming Li), (Corresponding authors: Yang Xiang; Dongning Sun; Zhiming Li)}}}

\markboth{Journal of \LaTeX\ Class Files,~Vol.~14, No.~8, August~2021}%
{Shell \MakeLowercase{\textit{et al.}}: A Sample Article Using IEEEtran.cls for IEEE Journals}


\maketitle

\begin{abstract}
Large language models (LLMs) have achieved substantial advances in logical reasoning, yet they continue to lag behind human-level performance. 
In-context learning provides a viable solution that boosts the model's performance via prompting its input with expert-curated, in-domain exemplars. However, in many real-world, expertise-scarce domains, such as low-resource scientific disciplines, emerging biomedical subfields, or niche legal jurisdictions, such high-quality in-domain demonstrations are inherently limited or entirely unavailable, thereby constraining the general applicability of these approaches.
To mitigate this limitation, recent efforts have explored the retrieval of cross-domain samples as surrogate in-context demonstrations. Nevertheless, the resulting gains remain modest. This is largely attributable to the pronounced domain shift between source and target distributions, which impedes the model’s ability to effectively identify and exploit underlying shared structures or latent reasoning patterns. Consequently, when relying solely on raw textual prompting, LLMs struggle to abstract and transfer such cross-domain knowledge in a robust and systematic manner. To address these issues, we propose \textbf{CoDA}, which employs a lightweight adapter to directly intervene in the intermediate hidden states. By combining feature-based distillation of CoT-enriched reference representations with Maximum Mean Discrepancy (MMD) for kernelized distribution matching, our method aligns the latent reasoning representation of the source and target domains. Extensive experimental results on multiple logical reasoning tasks across various model families validate the efficacy of CoDA by significantly outperforming the previous state-of-the-art baselines by a large margin.
\end{abstract}


\section{Introduction}

\IEEEPARstart{L}{}arge Language Models (LLMs) have achieved substantial advancements across various complex reasoning domains, including mathematical reasoning, programming, and biomedicine~\cite{zhao2023survey, imani2023mathprompter, huang2023towards}. These breakthroughs are largely driven by employing Chain-of-Thought (CoT) reasoning, which empowers models to decompose intricate problems into intermediate sub-tasks and resolve them step-by-step~\cite{wei2022chain}. Despite these advancements, they continue to lag behind human-level performance in systematic logical reasoning tasks that require rigorous step-by-step deduction to resolve complex multi-hop dependencies~\cite{JIANG2026108924, li2025investigatingtrainingdatadetection, Li_Jiang_Cao_Cui_Wu_Li_Liu_Sun_2025}. To improve reasoning performance, In-context learning provides a viable solution to boost the performance of the model by augmenting the input with expert-curated, in-domain exemplars~\cite{brown2020language, luo2024context, rubin2022learning, wang2024learning, ling2024uncertainty, lin2025reasoning}. However, in many real-world domains with scarce expertise, such as low-resource scientific disciplines, emerging biomedical subfields, or niche legal jurisdictions, high-quality in-domain demonstrations are inherently limited or entirely unavailable~\cite{li2024improving, Hu_Cao_Li_Li_Liu_Li_Chen_Liu_2024, 10.1145/3611643.3616351, pmlr-v202-li23h}. This critical data scarcity heavily constrains the broad applicability of such approaches in specialized scenarios. To mitigate this limitation, recent efforts have explored the retrieval of cross-domain samples to serve as surrogate in-context demonstrations~\cite{liu2026reasonanalogicallycrossdomainprior, tan2025shape}. For instance, approaches such as DIN-Retrieval attempt to extract universal hidden representations to fetch structurally compatible examples, leveraging the implicit logical patterns shared across disparate domains~\cite{yan2026effectiveincontextcrossdomainknowledge}. Nevertheless, the resulting performance gains remain notably modest. This bottleneck is largely attributable to the pronounced domain shift between the source and target distributions~\cite{tang2023large, sun2024exploring, siska2024examining}. Such divergence impedes the ability of the model to effectively identify and exploit underlying shared topological structures or latent reasoning patterns~\cite{besta2024graph, besta2025demystifying, bu2025enhanced}. Consequently, when relying solely on raw textual prompting, LLMs struggle to abstract and transfer cross-domain knowledge in a robust and systematic manner, necessitating deeper latent interventions to explicitly align these disparate representation spaces~\cite{huang2024unlocking, he2024using}.

To operationalize these deeper interventions, we introduce \underline{Co}T-guided \underline{D}omain \underline{A}daptation (\textbf{CoDA}), a framework that bypasses surface-level domain discrepancies by directly targeting shared reasoning patterns in the model's intermediate hidden states. Rather than relying on rigid projection matrices, CoDA employs a lightweight neural adapter to modulate internal representations. To achieve this, a Mean Squared Error (MSE) loss aligns latent source-domain reasoning states with Chain-of-Thought trajectories, while a Maximum Mean Discrepancy (MMD) loss extracts shared knowledge and reasoning patterns across domains~\cite{gretton2012kernel}. By explicitly aligning these latent reasoning representation, the method effectively bridges the semantic gap. It abstracts the underlying deduction logic from labeled source domains and seamlessly transfers it to unannotated target domains. This dynamic latent steering enables robust zero-shot inference in out-of-distribution scenarios, avoiding the mode collapse and overfitting typical of parametric updates. To validate the effectiveness of our framework, we comprehensively evaluate CoDA on cross-domain logical and mathematical reasoning tasks across various open-source architectures, spanning 12B to 32B parameter scales. Empirical results demonstrate that our latent intervention robustly transfers reasoning capabilities to unseen target domains, consistently outperforming existing baselines. In summary, our main contributions are threefold:
\begin{itemize}
\item \textbf{Identification of Latent Transfer Bottlenecks:} We empirically demonstrate that conventional text-level prompting and static activation interventions are insufficient for cross-domain CoT transfer, as they fail to capture and map the shared topological structures necessary for abstract reasoning.
\item \textbf{Novel Framework (CoDA):} We introduce Latent Representation Distillation, replacing rigid interventions with a lightweight neural adapter ($A_\theta$). This is coupled with a dual-objective loss—MSE for feature-based distillation and MMD for kernelized distribution alignment—to explicitly isolate and transfer domain-agnostic reasoning patterns.
\item \textbf{State-of-the-Art Zero-Shot Adaptation:} Our approach achieves substantial performance gains in zero-shot cross-domain reasoning, improving baseline accuracy by up to 12.3\% without requiring target-domain CoT annotations.
\end{itemize}

\section{Preliminaries}
\label{preliminaries}

In this section, we introduce the fundamentals of transfer learning, activation steering, and knowledge distillation.

\subsection{Domain Adaptation}
Domain adaptation aims to transfer knowledge from a labeled source domain to an unlabeled or under-resourced target domain whose data distribution differs from the source~\cite{pan2009survey, sun2015survey, farahani2021brief}. 
Let $\mathcal{D}_S = \{(x_i^S, y_i^S)\}_{i=1}^{n_S}$ denote the \textit{source domain} 
and $\mathcal{D}_T = \{(x_j^T)\}_{j=1}^{n_T}$ the \textit{target domain},
where $(x, y) \in \mathcal{X} \times \mathcal{Y}$.
Their input distributions differ:
\[
P_S(x) \neq P_T(x),
\]
while the underlying prediction function $f : \mathcal{X} \rightarrow \mathcal{Y}$ 
is assumed to be shared or related across domains.

The goal of domain adaptation is to find a mapping $f_\theta$ derived from $\mathcal{D}_S$ that generalizes to target samples $x^T \sim P_T(x)$ without access to labeled target data. A standard approach is to learn representations $h(x)\in\mathbb{R}^d$ that reduce the divergence between source and target feature distributions:
\[
\min_{\theta} \; 
\mathrm{Dist}\!\left(
    \{ h(x_i^{S}) \}_{i=1}^{n_S},
    \; 
    \{ h(x_j^{T}) \}_{j=1}^{n_T}
\right)
\]

where $\mathrm{Dist}(\cdot, \cdot)$ measures cross-domain divergence 

\subsection{Chain-of-Thought Reasoning}
Chain-of-Thought (CoT) reasoning is a prompting paradigm that empowers Large Language Models (LLMs) to decompose intricate, multi-step problems into a sequence of intermediate sub-tasks. By generating a series of intermediate reasoning steps, the model can resolve complex multi-hop dependencies that are otherwise difficult to capture via direct input-to-output mapping. Formally, let $\mathcal{X}$ denote the space of input queries and $\mathcal{Y}$ denote the space of final answers. In standard direct prompting, the model attempts to estimate the conditional probability $P(y|x)$. In the CoT framework, the model first generates a rationale or reasoning trajectory $c = (t_1, t_2, \dots, t_k) \in \mathcal{C}$, where $t_i$ represents intermediate tokens. The joint distribution of the rationale and the answer is formulated as:$$P(y, c | x) = P(c | x) \prod_{j=1}^{m} P(y_j | x, c, y_{<j})$$

where $P(c|x)$ characterizes the generation of the reasoning path. In supervised or few-shot scenarios, we define a source domain dataset $\mathcal{D}_S = \{(x_i^S, c_i^S, y_i^S)\}_{i=1}^{n_S}$, where each instance consists of a question $x_S$, an expert-curated CoT rationale $c_S$, and the ground-truth answer $y_S$.From a representational perspective, the inclusion of CoT rationales alters the internal latent states of the model. Given a feature extractor $f(\cdot)$ representing the initial $l$ layers of an LLM, the CoT-enriched "teacher" state $h^*$ can be extracted by processing the concatenation of the input and its rationale:$$\mathbf{h}^* = f(x \oplus c)$$While LLMs exhibit emergent reasoning capabilities, transferring these latent reasoning representation to unseen target domains $\mathcal{D}_T = \{x_j^T\}$ without explicit $c_T$ annotations remains a significant challenge due to pronounced domain shifts in the representation space.


\subsection{Chain-of-Thought Distillation}

Chain-of-Thought (CoT) distillation extends traditional knowledge distillation by transferring complex reasoning trajectories rather than merely matching final predictive logits. Standard text-based CoT distillation fine-tunes a student model $P_S(\cdot; \theta_S)$ by maximizing the log-likelihood of the teacher-generated rationale $c^*$ given an input $x$:

$$\mathcal{L}_{\text{Textual-CoT}} = - \sum_{t=1}^{|c^*|} \log P_S(c^*_t \mid x, c^*_{<t}; \theta_S)$$

To bypass the computational overhead of generating lengthy intermediate tokens and capture deeper topological structures, recent paradigms focus on \textit{latent representation distillation}. This approach directly aligns the student's intermediate hidden states $\mathbf{h}^S$ with the teacher's reasoning-enriched representations $\mathbf{h}^T$ via Mean Squared Error (MSE):
$$\mathcal{L}_{\text{Latent-CoT}} = \|\mathbf{h}^S(x) - \mathbf{h}^T(x \oplus c^*)\|_2^2$$
where $\oplus$ denotes sequence concatenation. By explicitly distilling these latent manifolds, models efficiently internalize structural deduction logic, establishing a robust foundation for cross-domain knowledge transfer.

\begin{figure*}[t!]
\centering
  \includegraphics[scale=0.3]{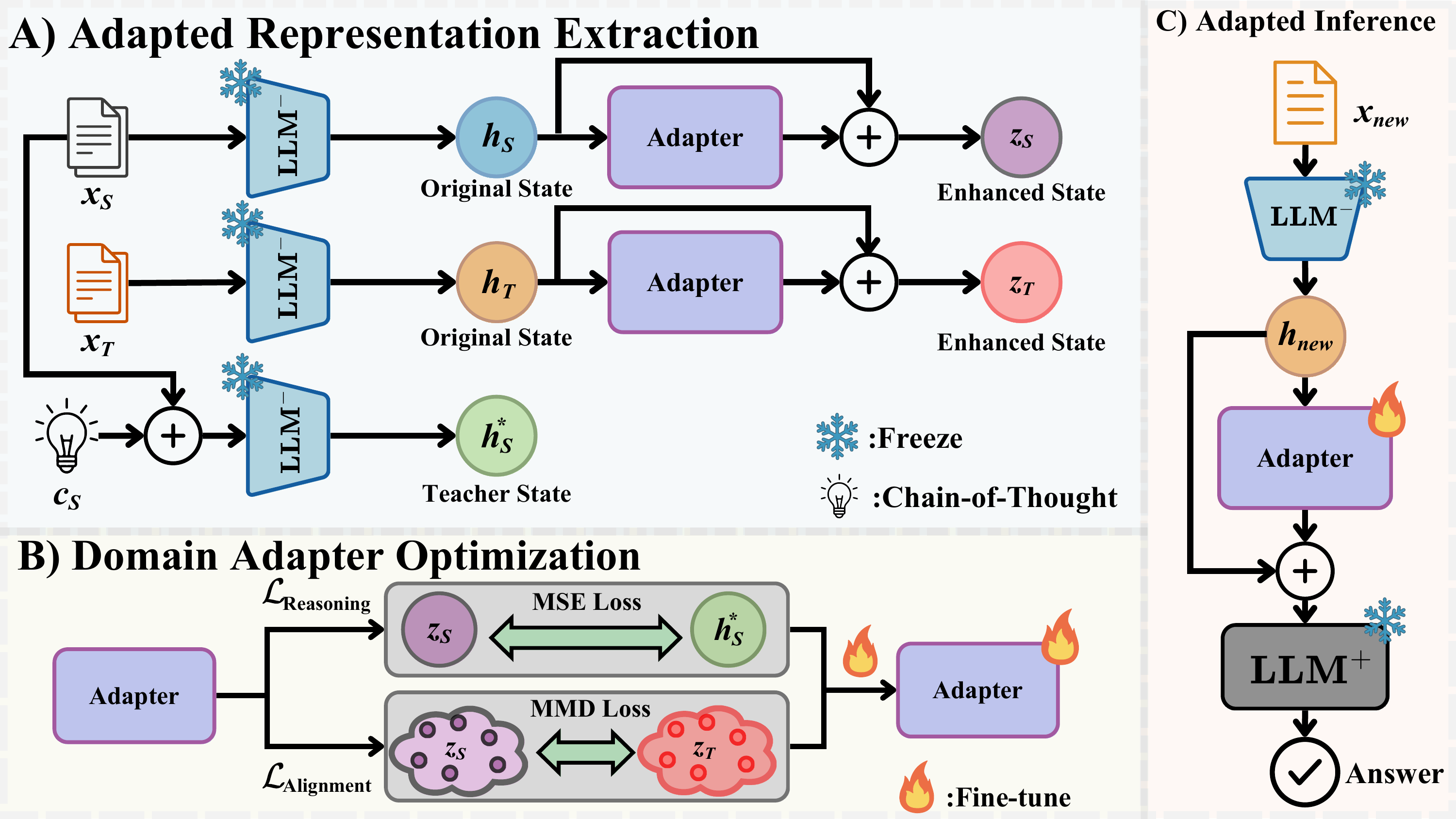}
  \caption{The overall framework of \textbf{CoDA}. (A) Adapted Representation Extraction: A frozen LLM extracts original hidden states ($h_S$, $h_T$) and the CoT-infused teacher state ($h_S^*$). A lightweight non-linear adapter is then employed to map these states into a shared reasoning-enhanced space. (B) CoT-guided Domain Adaptation: The adapter is optimized using a dual-objective: MSE loss distills the reasoning trajectory by minimizing the distance between $z_S$ and $h_S^*$ in the latent space such that the enhanced states approximate the teacher's reasoning state (i.e., $z_S \approx h_S^*$), while MMD loss aligns the source and target latent representation. (C) Adapted Inference: The optimized adapter directly modulates the hidden states of unseen target inputs, enabling robust zero-shot generation through the frozen reasoning generator.}
  \label{overview}
\end{figure*}

\section{Method}
\label{method}

As aforementioned, previous domain adaptation strategies rely on surface-level demonstrations, failing to bridge distribution shifts or align underlying reasoning representations essential for robust CoT transfer. To address this, we present CoDA. As shown in Figure~\ref{overview}, our framework aligns these latent reasoning spaces through three stages. First, to capture reasoning trajectories, it extracts standard hidden states alongside CoT-infused teacher states from a frozen language model. Next, to bridge the domain gap and facilitate transfer, we optimize a trainable adapter via a dual-objective strategy: feature-based alignment to mimic teacher representations and distribution alignment to merge source and target latent spaces. Finally, to enable robust zero-shot generation in unseen domains, this adapter directly intervenes in target hidden states during inference. The remainder of this section formalizes the problem setup, introduces latent intervention and extraction mechanisms, details dual-objective optimization, and outlines the zero-shot inference procedure.

\subsection{Adapted Representation Extraction}

The core of CoDA is to transfer reasoning capabilities from labeled source domain to unlabeled target domain directly via latent space intervention. We decompose a Large Language Model (LLM) into two sequential components: $M=LLM^- \circ LLM^+$, where $LLM^-(\cdot)$ extracts features from the first $l$ layers and $LLM^+(\cdot)$ generates reasoning from the remaining $L-l$ layers. Let $D_S=\{(x_S, c_S, y_S)\}$ be the source dataset, where each instance contains input question $x_S$, CoT rationale $c_S$, and answer $y_S$. Reflecting real-world unannotated environments, let $D_T=\{x_T\}$ be the target dataset containing only input questions without CoT annotations or labels. Finally, to bridge representational gaps via latent intervention, we introduce an adapter $A_\theta: \mathbb{R}^d \rightarrow \mathbb{R}^d$ parameterized by $\theta$. We implement $A_\theta$ via a residual connection to explicitly model the latent CoT shift~\cite{li2025cot}. Theoretically, including CoT rationales $c_S$ alters internal latent states by introducing new key-value dependencies in the self-attention mechanism. As shown in Part A of Figure~\ref{overview}, we first extract the original hidden states for both domains at the $l$-th layer:
\begin{equation}
    h_S=LLM^-(x_S), \quad h_T=LLM^-(x_T)
    \label{eq1}
\end{equation}

To facilitate cross-domain knowledge transfer, we also extract the reasoning representation of the labeled source domain data, which serves as a supervisory signal in the subsequent optimization stage. Formally, we construct a reference teacher state $h_S^*$ by passing the concatenation of the source input $x_S$ and its corresponding CoT rationale $c_S$ through the feature extractor $LLM^-(\cdot)$:

\begin{equation}
    h_S^*=LLM^-(x_S \oplus c_S)
    \label{eq2}
\end{equation}

where $\oplus$ denotes the sequence concatenation operator. Subsequently, we apply the non-linear adapter $A_\theta$ to inject reasoning capabilities into the original states via residual connections, yielding the enhanced states:

\begin{equation}
z_S=h_S+A_\theta(h_S),\quad z_T=h_T+A_\theta(h_T)
\label{eq3}
\end{equation}

\subsection{Domain Adapter Optimization}
\label{sec:objective}
As shown in Part B of Figure~\ref{overview}, to ensure the adapter $A_\theta$ learns robust and transferable reasoning representations, we jointly optimize two objectives during training. 

Reasoning Loss: On the source domain, we force the enhanced state $z_S$ to approximate the CoT-enriched reference state $h_S^*$. We minimize the Mean Squared Error (MSE) to distill the reasoning trajectory:

\begin{equation}
   \mathcal{L}_{Reason}(D_S;\theta)=||z_S-h_S^*||_2^2
\label{eq4}
\end{equation}

Distribution Alignment Loss: To mitigate domain shift and align the non-linear reasoning manifolds, we minimize the Maximum Mean Discrepancy (MMD)~\cite{gretton2012kernel} between the set of source enhanced states $Z_S=\{z_S^{(i)}\}_{i=1}^N$ and target enhanced states $Z_T=\{z_T^{(j)}\}_{j=1}^M$. Given a kernel function $k(\cdot, \cdot)$ mapping to a Reproducing Kernel Hilbert Space (RKHS), the empirical MMD is formulated as:

\begin{equation}
\label{eq5}
\begin{aligned}
\mathcal{L}_{MMD}^2(\mathcal{D}_S, \mathcal{D}_T) &= \underbrace{\frac{1}{N^2}\sum_{i,j} k(z_{S_i}, z_{S_j})}_{\text{Source Domain}} + \underbrace{\frac{1}{M^2}\sum_{i,j} k(z_{T_i}, z_{T_j})}_{\text{Target Domain}} \\
&\quad - \underbrace{\frac{2}{NM}\sum_{i,j} k(z_{S_i}, z_{T_j})}_{\text{Cross Domain}}
\end{aligned}
\end{equation}

The first two terms encourage intra-domain clustering 
, while the third term minimizes the cross-domain distance to ensure the target states closely resemble the source reasoning states.
The overall optimization objective is a weighted sum of the two losses:
\begin{equation}
    \min_\theta(L_{Reason}(D_S;\theta)+\lambda\cdot L_{MMD}(D_S,D_T;\theta))
\label{eq8}
\end{equation}

\subsection{Adapted Inference}
Finally, in Part C of Figure~\ref{overview}, during inference, for any new sample $x_{new} \in D_T$, we freeze the base LLM and only utilize the trained adapter to perform zero-shot reasoning. The process is defined as: \\
1. Extract the initial state:  \\
\begin{equation}
    h_{new}=LLM^-(x_{new})
\end{equation}
2. Apply activation steering: \\
\begin{equation}
    z_{new}=h_{new}+A_\theta(h_{new})
\label{eq10}
\end{equation}
3. Generate the final output: \\
\begin{equation}
    \hat{g}=LLM^+(z_{new})
\label{eq11}
\end{equation}

As shown in Algorithm~\ref{alg:coda}, this plug-and-play steering mechanism empowers the frozen LLM to generate logical trajectories for out-of-domain queries efficiently and effectively.

\begin{algorithm}[t]
\caption{CoDA: CoT-guided Domain Adaptation}
\label{alg:coda}
\begin{algorithmic}[1]
    \renewcommand{\algorithmicrequire}{\textbf{Input:}}
    \renewcommand{\algorithmicensure}{\textbf{Output:}}
    \REQUIRE Source $\mathcal{D}_S = \{(x_S, c_S, y_S)\}$, Target $\mathcal{D}_T = \{x_T\}$; \\
             LLM components $\{LLM^-(\cdot), LLM^+(\cdot)\}$; Strength $\alpha$, Penalty $\lambda$.
    \ENSURE Optimized adapter $A_\theta$.
    
    \vspace{0.5em}
    \STATE \textit{// Phase 1: Domain Adapter Optimization}
    \WHILE{not converged}
        \STATE Sample batches $\{(x_S, c_S)\}_{i=1}^N \subset \mathcal{D}_S$ and $\{x_T\}_{j=1}^M \subset \mathcal{D}_T$
        
        \STATE \COMMENT{\textcolor{gray}{Adapted Representation Extraction \& Steering}}
        \STATE Compute $h_S, h_T = LLM^-(x_S), LLM^-(x_T)$ \hfill $\triangleright$ Eq.~\ref{eq1}
        \STATE Compute teacher state $h_S^* = f(x_S \oplus c_S)$ \hfill $\triangleright$ Eq.~\ref{eq2}
        \STATE Obtain enhanced states $z = h + A_\theta(h)$ for $S$ and $T$ \\ \hfill $\triangleright$ Eq.~\ref{eq3}
        
        \STATE \COMMENT{\textcolor{gray}{Joint Objective Optimization}}
        \STATE $\mathcal{L}_{Reason} = \|z_S - h_S^*\|_2^2$ \hfill $\triangleright$ Eq.~\ref{eq4}
        \STATE $\mathcal{L}_{MMD} = \text{MMD}(\mathcal{Z}_S, \mathcal{Z}_T)$ \hfill $\triangleright$ Eq.~\ref{eq5}
        \STATE Update $\theta$ by minimizing $\mathcal{L} = \mathcal{L}_{Reason} + \lambda \cdot \mathcal{L}_{MMD}$ \\ \hfill $\triangleright$ Eq.~\ref{eq8}
    \ENDWHILE

    \vspace{0.5em}
    \STATE \textit{// Phase 2: Adapted Inference}
    \STATE \textbf{Input:} New sample $x_{new} \in D_T$
    \STATE 1. Apply latent steering: \\ $z_{new} = LLM^-(x_{new}) + A_\theta(LLM^-(x_{new}))$ \hfill $\triangleright$ Eq.~\ref{eq10}
    \STATE 2. Generate final reasoning: $\hat{g} = LLM^+(z_{new})$ \hfill $\triangleright$ Eq.~\ref{eq11}
    \STATE \textbf{return} $\hat{g}$
\end{algorithmic}
\end{algorithm}

\begin{table*}[t]
\centering
\small
\caption{Task categorization and datasets used for cross-domain reasoning evaluation.}
\begin{tabular}{lllcc}
\toprule
\textbf{Task Type} & \textbf{Dataset Name} &  \textbf{Description}& \textbf{Train Data Size} & \textbf{Test Data Size} \\
\midrule
\multirow{2}{*}{\makecell{Mathematical reasoning}} 
     & \multirow{2}{*}{\makecell{GSM8K}~\cite{cobbe2021training}} & \multirow{2}{*}{\makecell{Math Problems}}  & \multirow{2}{*}{\makecell{7037}} & \multirow{2}{*}{\makecell{1319}} \\
    \\
\midrule
\multirow{3}{*}{\makecell{Logical Reasoning}} 
     & LogicalDeduction~\cite{nguyen2025non} & Deductive logical reasoning & 1200 & 300 \\
     & FOLIO~\cite{han2024folio} & First-order logic reasoning & 1004 & 204 \\
     & ProofWriter~\cite{tafjord2021proofwriter} & Generating implications & 3000 & 600 \\
\midrule
\multirow{2}{*}{\makecell{CommonSense}} 
     & \multirow{2}{*}{CommonSenseQA~\cite{talmor2019commonsenseqa}} & \multirow{2}{*}{\makecell{Broad commonsense reasoning}} & \multirow{2}{*}{9741} & \multirow{2}{*}{1221} \\
    \\
\bottomrule
\end{tabular}
\label{tab:task-type-domain}
\vspace{-0.2in}
\end{table*}

\section{Experiments}

In this section, we validate the effectiveness of \AdaSteer in terms of cross-domain knowledge transfer by addressing three core research questions:
\textbf{RQ1} — How effective is \AdaSteer regarding cross-domain knowledge transfer compared to the state-of-the-art baselines? (Section~\ref{rq1})
\textbf{RQ2} — Does the \AdaSteer effectively align cross-domain reasoning representation? (Section~\ref{rq2})
\textbf{RQ3} — How parameter-efficient is \AdaSteer?
\textbf{RQ4} — How do the key components of \AdaSteer impact overall transfer effectiveness? (Section~\ref{rq3})

\subsection{Backbone Model}
We evaluate \AdaSteer across a diverse set of open-source large language models to ensure our findings generalize across different architectures and parameter scales. Our experiments utilize \textbf{Gemma-3-it} (4B, 12B and 27B) \cite{gemmateam2025gemma3technicalreport}, and the \textbf{Qwen-2.5-Instruct} (3B, 7B, 14B and 32B) \cite{qwen2.5}. By encompassing both moderate and larger capacities, we thoroughly demonstrate the scalability of our approach.

\subsection{Datasets \& Tasks}

As shown in Table~\ref{tab:task-type-domain}, to evaluate the effectiveness of CoDA regarding cross-domain knowledge transfer of \AdaSteer, we utilize four diverse datasets spanning mathematical and logical reasoning domains. For the source domain, each instance consists of an input question $x_S$, a Chain-of-Thought (CoT) rationale $c_S$, and a final answer $y_S$. In contrast, the target domain datasets are treated as unlabeled, containing only input questions $x_T$ without any CoT annotations or labels during training.
\begin{itemize}
    \item \textbf{GSM8K}~\cite{cobbe2021training}: This dataset consists of high-quality grade school math word problems that require multi-step reasoning to solve. In our experiments, it serves as the primary representative for the mathematical reasoning domain.
    \item \textbf{LogicalDeduction}~\cite{nguyen2025non}: This dataset is used to test deductive reasoning capabilities, where the model must infer conclusions from a set of provided logical premises.
    \item \textbf{FOLIO}~\cite{han2024folio}: A natural language reasoning dataset based on first-order logic. It requires the model to determine whether a conclusion is True, False, or Uncertain based on complex logical constraints.
    \item \textbf{ProofWriter}~\cite{tafjord2021proofwriter}: This dataset focuses on generating implications and proofs over natural language. It challenges the model to perform multi-step reasoning to derive implications from a given theory.
    \item \textbf{CommonSenseQA}~\cite{talmor2019commonsenseqa}: A multiple-choice question answering dataset designed to evaluate broad commonsense reasoning capabilities over everyday background knowledge.
\end{itemize}

\begin{table*}[t!]
\centering
\small
\caption{Comparison of evaluated baselines across different paradigms. We systematically categorize methods based on their trainability, intervention space, and core mechanisms.}
\begin{tabular}{llcll}
\toprule
\textbf{Paradigm} & \textbf{Method} & \textbf{Trainable} & \textbf{Intervention Space} & \textbf{Description} \\
\midrule
\multirow{3}{*}{Retrieval} 
    & BM25~\cite{robertson2009probabilistic}   & \ding{56}  & Context & Sparse text retrieval \\
    & Embed~\cite{lewis2020retrieval} & \ding{56}  & Context & Dense vector retrieval \\
    & ConE~\cite{peng2024revisiting}   & \ding{56}  & Context & Perplexity-based Rerank \\
    & DIN~\cite{yan2026effectiveincontextcrossdomainknowledge} & \ding{56}  & Context & Universal-neurons-based Retrieval \\
\midrule
\multirow{2}{*}{PEFT} 
    & LoRA~\cite{hu2022lora}       & \ding{52} & Parameters & Low-rank adaptation \\
    & P-tuning~\cite{liu2022p}& \ding{52} & Input Embeddings & Continuous prompt optimization \\
\midrule
\multirow{3}{*}{Steering} 
    & CAA~\cite{panickssery2023steering}               & \ding{56} & Activations & Linear activation steering \\
    & CoT-Vectors~\cite{li2025cot} & \ding{52} & Activations & Trainable vector steering \\
    & \textbf{\AdaSteer (Ours)}     & \ding{52} & \textbf{Activations} & \textbf{Adaptation-based steering} \\
\bottomrule
\end{tabular}
\label{tab:baselines_summary}
\vspace{-0.15in}
\end{table*}

\begin{table}[t!]
\centering
\small
\caption{Implementation details and hyperparameter settings.}
\setlength{\tabcolsep}{4pt}
\begin{tabular}{llc}
\toprule
\textbf{Method} & \textbf{Hyperparameter} & \textbf{Value} \\
\midrule
\multirow{9}{*}{\textbf{\AdaSteer}} 
    & Steering Layer ($l$) & 15 $\sim$ 34 \\
    & Steering Strength ($\alpha$) & 1.5 \\
    & MMD Penalty ($\lambda$) & 1.0 \\
    & Learning Rate & $1e^{-4}$ \\
    & Train Batch Size & 32 \\
    & Eval Batch Size & 8 \\
    & Max Epochs & 50 \\
    & Optimizer & Adam~\cite{diederik2014adam} \\
    & Initialization & Zero \\
\midrule
\multirow{5}{*}{\textbf{LoRA}} 
    & Rank ($r$) &  8, 16 \\
    & Alpha ($\alpha$) & 16 \\
    & Target Modules & All \\
    & Dropout & 0.05 \\
    & Initialization & Zero \\
\midrule
\multirow{4}{*}{\textbf{P-tuning}} 
    & Learning Rate & $1e^{-4}$ \\
    & Train Batch Size & 32 \\
    & Initialization & Zero \\
    & Max Epochs & 50 \\
\midrule
\multirow{2}{*}{\textbf{CAA}} 
    & Steering Layer ($l$) & 15 $\sim$ 34 \\
    & Vector Normalization & Across behaviors \\
\midrule
\multirow{4}{*}{\textbf{CoT-Vectors}} 
    & Steering Layer ($l$) & 15 $\sim$ 34 \\ 
    & Loss factor ($\lambda$) & 0.5 \\
    & Steering Strength ($\alpha$) & 1.0 \\
    & Optimizer & Adam~\cite{diederik2014adam} \\
\midrule
\multirow{2}{*}{\textbf{Embedding}} 
    & Embedding Model & bge-large-en-v1.5~\cite{xiao2024c} \\
    & Top-k & 4 \\
\midrule
\multirow{2}{*}{\textbf{BM25}} 
    & Term Frequency ($k_1$) & 1.5 \\
    & Length Normalization ($b$) & 0.75 \\
\midrule
\multirow{2}{*}{\textbf{ConE}} 
    & Top-k & 30 \\
    & Number of Examples ($N$) & 4 \\
\bottomrule
\label{tab:hyperparameters}
\end{tabular}

\vspace{-0.15in}
\end{table}

\subsection{Baselines}
To benchmark the effectiveness of \AdaSteer, we compare our framework against the standard zero-shot baseline, alongside recent and representative methods that can be directly/indirectly leveraged for cross-domain knowledge transfer, namely retrieval (which boost the model's prompting with cross-domain exemplars that are of similar task-solving logic), parameter-efficient fine-tuning (PEFT) (optimizes lightweight source-domain parameters encoding universal patterns and transferring to target domain), and steering (isolating reasoning-specific source activation directions to dynamically modulate intermediate target representations). We provide a concise introduction of the baseline methods as below (refer to Table~\ref{tab:baselines_summary} for the parallel comparison and  Table~\ref{tab:hyperparameters} for the implementation details):

\begin{itemize}
    \item \textbf{Zero-shot:} Evaluates the base model's direct reasoning capabilities without providing any external context, demonstrations, or parameter updates.
    
    \item \textbf{BM25}~\cite{robertson2009probabilistic}: A non-trainable sparse text retrieval method that fetches relevant context based on exact keyword matching.
    
    \item \textbf{Embed}~\cite{lewis2020retrieval}: A dense vector retrieval baseline that fetches semantically relevant context using neural text embeddings.
    
    \item \textbf{ConE}~\cite{peng2024revisiting}: A context intervention method that utilizes perplexity-based reranking to select the most helpful retrieved examples.
    
    \item \textbf{LoRA}~\cite{hu2022lora}: A parameter-efficient fine-tuning (PEFT) approach that adapts the model to target domains by optimizing injected low-rank matrices.
    
    \item \textbf{P-tuning}~\cite{liu2022p}: A trainable baseline that performs continuous prompt optimization directly on the input embeddings.
    
    \item \textbf{Contrastive Activation Addition (CAA)}~\cite{panickssery2023steering}: A non-trainable method that manipulates reasoning via contrastive linear vectors. We extract a steering vector (the difference between source CoT and standard states) and inject it into target activations during inference.
    
    \item \textbf{CoT-Vectors}~\cite{li2025cot}: A trainable activation steering approach that employs optimized vectors to intervene in the model's internal reasoning pathways by adding a steering vector, optimized on the source domain, to the hidden activations during target-domain inference.
\end{itemize}

\begin{table}[t!]
\centering
\small
\caption{Implementation details and hyperparameter settings.}
\begin{tabular}{lcc}
\toprule
\textbf{Backbone Model} & \textbf{Steering Layer ($l$)} & \textbf{Total Layers} \\
\midrule
Qwen-2.5-3B-Instruct  & 15 & 36 \\
Qwen-2.5-7B-Instruct  & 14 & 28 \\
Qwen-2.5-14B-Instruct & 24 & 48 \\
Qwen-2.5-32B-Instruct & 34 & 64 \\
\midrule
Gemma-3-4B-it  & 17 & 34 \\
Gemma-3-12B-it & 24 & 48 \\
Gemma-3-27B-it & 34 & 62 \\
\midrule
Llama-3.1-8B-Instruct & 16 & 32 \\
\bottomrule
\end{tabular}
\label{tab:steerlayer}
\vspace{-0.15in}
\end{table}

\subsection{Implementation details}

\paragraph{Hyperparameters}
We summarize the comprehensive hyperparameter configurations for \AdaSteer and all evaluated baselines in Table~\ref{tab:hyperparameters}. For our proposed framework, the non-linear adapter is optimized using the Adam optimizer~\cite{diederik2014adam} with a learning rate of $1 \times 10^{-4}$ and a training batch size of 32 for a maximum of 50 epochs. Based on our empirical sensitivity analysis, we set the default steering strength to $\alpha = 1.5$ and the MMD alignment penalty to $\lambda = 1.0$. 

For all activation steering methods (including \AdaSteer, CAA, and CoT-Vectors), the optimal intervention layer $l$ is highly dependent on the underlying model architecture. We consistently conduct interventions across the mid-to-late transformer blocks (ranging from layer 17 to 34). The exact layer selections corresponding to each backbone model are detailed in Table~\ref{tab:steerlayer}.

To ensure a fair comparison, the parameter-efficient fine-tuning (PEFT) baselines (LoRA and P-tuning) share the identical optimization setup (learning rate and batch size) with \AdaSteer. For LoRA, we apply low-rank updates to all linear target modules with a rank $r \in \{8, 16\}$, a scaling factor $\alpha = 16$, and a dropout rate of $0.05$~\footnote{We use the LlamaFactory framework~\cite{zheng2024llamafactory} for training}. For retrieval-augmented baselines, we set the number of retrieved in-context demonstrations to $k=4$, utilizing \texttt{bge-large-en-v1.5} for dense embedding retrieval and standard penalty parameters ($k_1=1.5, b=0.75$) for sparse BM25 retrieval.

\subsection{Cross-Domain Knowledge Transfer Effectiveness (RQ1)}
\label{rq1}

\begin{table*}[ht!]
\centering
\small
\caption{\textbf{Comprehensive cross-domain reasoning results.} \AdaSteer consistently outperforms Retrieval, PEFT, and alternative Steering methods across various source-to-target domain pairs. The table includes both main backbones (14B/32B/12B/27B) and additional smaller variants (3B/4B/7B/8B). Highlighting \colorbox{boldblue}{best} among same-sized models. Avg. denotes the average score across all tasks.}
\resizebox{\textwidth}{!}{
\begin{tabular}{l ccccccccc ccccccccc}
\toprule 
& \multicolumn{8}{c}{\textbf{Source Domain $\rightarrow$ Target Domain}} &  & \multicolumn{8}{c}{\textbf{Source Domain $\rightarrow$ Target Domain}} &  \\
\cmidrule(lr){2-10} \cmidrule(lr){11-19}

& \multicolumn{9}{c}{\textbf{Qwen-2.5-3B-Instruct}} & \multicolumn{9}{c}{\textbf{Gemma-3-4B-it}} \\

\cmidrule(lr){2-10} \cmidrule(lr){11-19}
\textbf{Method} & \textit{P-L} & \textit{L-P} & \textit{F-L} & \textit{L-F} & \textit{P-F} & \textit{F-P} & \textit{G-L} & \textit{C-L} & \avgbg Avg. & \textit{P-L} & \textit{L-P} & \textit{F-L} & \textit{L-F} & \textit{P-F} & \textit{F-P} & \textit{G-L} & \textit{C-L} & \avgbg Avg. \\
\midrule

Zero-shot                & 41.7 & 52.8 & 41.7 & 63.2 & 63.2 & 52.8 & 41.7 & 41.7 & \avgbg 49.9 & 60.7 & 62.7 & 60.7 & 61.8 & 61.8 & 62.7 & 60.7 & 60.7 & \avgbg 61.5 \\
Embed                    & 45.3 & 54.3 & 45.3 & 60.8 & 64.2 & 53.7 & 46.3 & 44.3 & \avgbg 51.8 & 55.3 & 54.8 & 58.7 & 63.2 & 61.8 & 63.2 & 56.3 & 54.3 & \avgbg 58.5 \\
BM25                     & 46.0 & 54.3 & 44.0 & 59.8 & 61.3 & 54.3 & 47.0 & 45.0 & \avgbg 51.5 & 41.3 & 57.0 & 41.0 & 63.7 & 66.7 & 61.5 & 42.3 & 40.3 & \avgbg 51.7 \\
ConE                     & 46.7 & 55.7 & 46.7 & 63.7 & 64.2 & 55.5 & 47.7 & 45.7 & \avgbg 53.2 & 58.7 & 53.2 & 62.7 & 61.8 & 64.2 & 61.8 & 59.7 & 57.7 & \avgbg 60.0 \\
LoRA                     & 43.2 & 42.0 & 42.6 & 51.0 & 53.7 & 43.0 & 44.2 & 42.2 & \avgbg 45.2 & 60.2 & 41.1 & 60.2 & 49.5 & 50.4 & 42.5 & 61.2 & 59.2 & \avgbg 53.0 \\
P-tuning & 45.3 & 43.8 & 44.8 & 52.9 & 55.7 & 45.1 & 45.9 & 44.4 & \avgbg 47.2 & 62.2 & 43.0 & 62.3 & 51.5 & 52.2 & 44.7 & 63.1 & 61.3 & \avgbg 55.0 \\
CAA                      & 47.6 & 53.9 & 46.6 & 61.3 & 61.9 & 53.2 & 48.6 & 46.6 & \avgbg 52.5 & 66.2 & 62.2 & 68.8 & 58.4 & 60.5 & 64.3 & 67.2 & 65.2 & \avgbg 64.1 \\
CoT-Vectors              & 48.7 & 55.0 & 47.7 & 62.4 & 63.0 & 54.3 & 49.7 & 47.7 & \avgbg 53.6 & 67.1 & \bc 63.4 & 69.9 & 59.4 & 61.3 & \bc 65.4 & 68.1 & 66.4 & \avgbg 65.1 \\
DIN                      & 49.2 & 57.8 & 49.4 & 63.7 & \bc 66.4 & 55.5 & 52.7 & \bc 52.6 & \avgbg 55.9 & 71.0 & 56.0 & 71.5 & 66.0 & 67.5 & 57.5 & 72.0 & 70.0 & \avgbg 66.4 \\
\AdaSteer                & \bc 55.3 & \bc 61.3 & \bc 52.3 & \bc 64.3 & 64.3 & \bc 56.2 & \bc 56.3 & 54.3 & \avgbg \bc \bf 58.0 & \bc 73.0 & 57.5 & \bc 72.7 & \bc 67.6 & \bc 68.1 & 58.0 & \bc 74.0 & \bc 72.0 & \avgbg \bc \bf 67.9 \\

\midrule
& \multicolumn{9}{c}{\textbf{Qwen-2.5-7B-Instruct}} & \multicolumn{9}{c}{\textbf{Llama-3.1-8B-Instruct}} \\
\midrule

Zero-shot                & 62.0 & 62.5 & 62.0 & 69.6 & 69.6 & 62.5 & 63.0 & 63.0 & \avgbg 64.0 & 61.5 & 63.0 & 62.8 & 68.5 & 70.2 & 61.8 & 62.0 & 62.0 & \avgbg 64.2 \\
Embed                    & 63.7 & 62.0 & 63.3 & 68.6 & 70.1 & 61.3 & 64.7 & 62.7 & \avgbg 64.6 & 64.2 & 61.5 & 64.0 & 69.0 & 69.5 & 62.1 & 65.3 & 61.8 & \avgbg 64.7 \\
BM25                     & 60.7 & 61.2 & 65.7 & 69.1 & 71.1 & 61.7 & 61.7 & 59.7 & \avgbg 63.9 & 61.2 & 60.8 & 64.9 & 68.5 & 71.5 & 62.4 & 62.3 & 60.5 & \avgbg 64.0 \\
ConE                     & 65.0 & 61.3 & 68.3 & 68.1 & 70.1 & 65.0 & 66.0 & 64.0 & \avgbg 66.0 & 64.5 & 62.0 & 67.5 & 68.8 & 71.0 & 64.3 & 65.2 & 64.8 & \avgbg 66.0 \\
LoRA                     & 59.2 & 47.4 & 61.1 & 57.0 & 60.0 & 49.6 & 60.2 & 58.2 & \avgbg 56.6 & 58.5 & 48.2 & 62.0 & 58.1 & 59.5 & 50.4 & 59.5 & 57.6 & \avgbg 56.7 \\
P-tuning & 61.1 & 49.5 & 63.1 & 58.8 & 62.2 & 51.5 & 62.3 & 60.2 & \avgbg 58.6 & 60.6 & 50.0 & 64.0 & 60.0 & 61.6 & 52.6 & 61.3 & 59.6 & \avgbg 58.7 \\
CAA                      & 65.2 & 60.9 & 66.7 & 68.4 & 69.2 & 61.4 & 66.2 & 64.2 & \avgbg 65.3 & 66.0 & 61.5 & 67.1 & 67.8 & 70.0 & 62.2 & 65.5 & 65.1 & \avgbg 65.7 \\
CoT-Vectors              & 66.3 & 62.0 & 67.8 & 69.5 & 70.3 & 62.5 & 67.3 & 65.3 & \avgbg 66.4 & 67.2 & 62.4 & 68.2 & 68.6 & 71.3 & 63.2 & 66.6 & 66.0 & \avgbg 66.7 \\
DIN                      & 67.5 & \bc 65.6 & 69.4 & 71.2 & 72.0 & 67.5 & 68.0 & 68.0 & \avgbg 68.8 & 69.5 & 62.0 & 68.5 & 69.0 & 71.0 & 65.5 & 69.5 & 68.5 & \avgbg 67.9 \\
\AdaSteer                & \bc 70.3 & 64.6 & \bc 70.0 & \bc 73.2 & \bc 73.7 & \bc 67.5 & \bc 71.3 & \bc 69.3 & \avgbg \bc \bf 70.0 & \bc 71.0 & \bc 63.5 & \bc 69.5 & \bc 70.2 & \bc 71.8 & \bc 66.1 & \bc 70.8 & \bc 70.1 & \avgbg \bc \bf 69.1 \\

\midrule
& \multicolumn{9}{c}{\textbf{Qwen-2.5-14B-Instruct}} & \multicolumn{9}{c}{\textbf{Gemma-3-12B-it}} \\
\midrule

Zero-shot                & 79.7 & 67.7 & 79.7 & 68.1 & 68.1 & 67.7 & 79.7 & 79.7 & \avgbg 73.7 & 73.3 & 73.8 & 73.3 & 71.1 & 71.1 & 73.8 & 73.7 & 73.7 & \avgbg 73.6 \\
Embed                    & 79.7 & 65.7 & 80.0 & 74.0 & 72.1 & 65.0 & 80.2 & 79.8 & \avgbg 74.5 & 78.3 & 76.2 & 76.0 & 72.5 & 73.5 & 76.0 & 78.5 & 79.0 & \avgbg 76.2 \\
BM25                     & 78.7 & 69.5 & 78.3 & 72.5 & 74.0 & 70.8 & 81.0 & 80.5 & \avgbg 75.6 & 65.3 & 75.7 & 66.7 & 71.1 & 75.5 & 74.5 & 76.5 & 75.8 & \avgbg 72.6 \\
ConE                     & 80.3 & 69.8 & 79.3 & 72.5 & 73.5 & 71.5 & 82.1 & 81.7 & \avgbg 76.3 & 77.3 & 76.8 & 85.7 & 73.5 & 76.5 & 75.3 & 81.2 & 81.5 & \avgbg 78.4 \\
LoRA                     & 74.0 & 52.0 & 74.0 & 60.0 & 62.0 & 54.0 & 69.5 & 71.0 & \avgbg 64.1 & 75.0 & 53.0 & 75.0 & 62.7 & 63.7 & 54.6 & 68.5 & 67.0 & \avgbg 64.9 \\
P-tuning & 76.0 & 54.1 & 75.9 & 62.0 & 63.8 & 56.2 & 71.5 & 73.1 & \avgbg 66.6 & 76.8 & 55.0 & 77.1 & 64.6 & 65.7 & 56.8 & 70.4 & 69.1 & \avgbg 66.9 \\
CAA                      & 81.5 & 66.8 & 80.8 & 72.0 & 71.5 & 66.8 & 81.0 & 81.5 & \avgbg 75.2 & 82.0 & 77.3 & 85.0 & 73.0 & 75.4 & 79.8 & 82.4 & 82.0 & \avgbg 79.6 \\
CoT-Vectors              & 82.6 & 67.9 & 81.5 & 73.1 & 72.6 & 67.9 & 82.1 & 82.4 & \avgbg 76.3 & 83.2 & 78.3 & 85.8 & 74.1 & 76.6 & 80.4 & 83.5 & 83.3 & \avgbg 80.7 \\
DIN                      & 80.8 & \bc 71.4 & 79.2 & 75.9 & 75.9 & \bc 72.9 & 81.8 & 82.5 & \avgbg 77.5 & 83.5 & 77.0 & 84.5 & 73.5 & 76.5 & 79.5 & 83.0 & 83.0 & \avgbg 80.1 \\
\AdaSteer                & \bc 83.6 & 70.1 & \bc 81.6 & \bc 76.4 & \bc 75.9 & 72.1 & \bc 82.8 & \bc 82.5 & \avgbg \bc \bf 78.1 & \bc 85.0 & \bc 78.5 & \bc 86.0 & \bc 74.5 & \bc 77.7 & \bc 80.5 & \bc 84.2 & \bc 84.3 & \avgbg \bc \bf 81.3 \\

\midrule
& \multicolumn{9}{c}{\textbf{Qwen-2.5-32B-Instruct}} & \multicolumn{9}{c}{\textbf{Gemma-3-27B-it}} \\
\midrule

Zero-shot                & 83.0 & 68.8 & 83.0 & 72.5 & 72.5 & 68.8 & 83.0 & 83.0 & \avgbg 76.7 & 77.7 & 77.3 & 77.7 & 71.1 & 71.1 & 77.3 & 77.7 & 77.7 & \avgbg 77.2 \\
Embed                    & 84.0 & 69.7 & 83.7 & 72.5 & 73.0 & 70.2 & 84.2 & 83.9 & \avgbg 77.6 & 87.3 & 76.7 & 82.7 & 72.1 & 74.0 & 79.8 & 86.5 & 86.2 & \avgbg 80.6 \\
BM25                     & 86.7 & 70.2 & 83.7 & 72.1 & 72.5 & 71.3 & 85.1 & 84.5 & \avgbg 78.2 & 85.3 & 78.5 & 84.3 & 71.6 & 76.0 & 78.2 & 85.8 & 85.5 & \avgbg 80.6 \\
ConE                     & 86.0 & \bc 72.5 & 84.7 & 74.5 & 73.5 & \bc 73.5 & 85.8 & 85.6 & \avgbg 79.5 & 85.5 & 79.2 & 85.7 & 75.5 & 76.5 & 80.8 & 88.5 & 88.0 & \avgbg 82.4 \\
LoRA                     & 76.0 & 70.0 & 78.0 & 73.0 & 74.0 & 70.0 & 78.5 & 79.2 & \avgbg 74.8 & 72.3 & 76.5 & 81.0 & 75.0 & 77.9 & 73.3 & 84.5 & 83.5 & \avgbg 78.0 \\
P-tuning & 78.0 & 71.9 & 80.1 & 75.0 & 75.8 & 72.2 & 80.4 & 81.3 & \avgbg 76.8 & 74.4 & 78.3 & 83.0 & 76.9 & 80.0 & 75.3 & 86.3 & 85.7 & \avgbg 80.0 \\
CAA                      & 85.5 & 70.2 & 86.0 & 74.5 & 74.0 & 70.2 & 85.3 & 84.8 & \avgbg 78.8 & 87.7 & 73.5 & 90.3 & 72.5 & 78.6 & 70.1 & 87.2 & 86.8 & \avgbg 80.8 \\
CoT-Vectors              & 86.6 & 71.2 & 87.2 & 75.6 & 74.9 & 71.3 & 86.5 & 85.9 & \avgbg 79.9 & 88.8 & 74.7 & \bc 91.4 & 73.4 & \bc 79.5 & 71.2 & 88.1 & 88.0 & \avgbg 81.9 \\
DIN                      & 86.5 & 71.0 & 87.0 & 75.5 & 75.0 & 71.0 & 86.0 & 85.5 & \avgbg 79.7 & 90.0 & 80.0 & 85.5 & 75.5 & 78.0 & 81.5 & 89.5 & 89.5 & \avgbg 83.7 \\
\AdaSteer                & \bc 88.0 & 71.6 & \bc 88.3 & \bc 76.4 & \bc 75.9 & 71.8 & \bc 87.0 & \bc 86.6 & \avgbg \bc \bf 80.7 & \bc 91.6 & \bc 81.0 & 87.0 & \bc 76.4 & 79.1 & \bc 82.8 & \bc 91.2 & \bc 91.0 & \avgbg \bc \bf 85.0 \\

\bottomrule
\end{tabular}
}
\label{tab:main}
\end{table*}

To evaluate the effectiveness of \AdaSteer regarding cross-domain knowledge transfer, we compared three representative baselines across 8 transfer directions. As shown in Table~\ref{tab:main}, \AdaSteer outperforms existing methods across all evaluated architectures and scales. Due to the validity of our proposed CoT-guided domain adapter, our approach surpasses traditional retrieval-augmented in-context learning and parameter-efficient fine-tuning (PEFT) by a large margin. For instance, on Gemma-3-it 27B, \AdaSteer achieves an average accuracy of 85.0\%, outperforming the strongest retrieval baseline ConE (82.4\%) and exceeding zero-shot performance by 7.8\%. In contrast, standard PEFT methods like LoRA and P-tuning frequently underperform zero-shot baseline due to severe source distribution overfitting. On the Qwen-2.5-Instruct 14B backbone, LoRA yields an average of 64.1\%, substantially lower than the 73.7\% achieved by zero-shot inference, indicating that parametric updates alone are prone to mode collapse. Compared to activation steering, \AdaSteer demonstrates higher stability. While Contrastive Activation Addition remains competitive, the trainable CoT-Vectors method experiences a severe performance drop, yielding 27.7\% on Gemma-3-it 27B. The results validate \AdaSteer's superiority compared to the baseline methods in terms of cross-domain knowledge transfer.

\begin{figure*}[h!]
    \caption{Illustrative reasoning trajectories. Compared to baselines where zero-shot inference hallucinates loops, LoRA exhibits mode collapse, and DIN suffers from semantic misalignment, \AdaSteer maintains structural coherence by directly identifying core contradictions.}
  \centering
  \includegraphics[width=\linewidth]{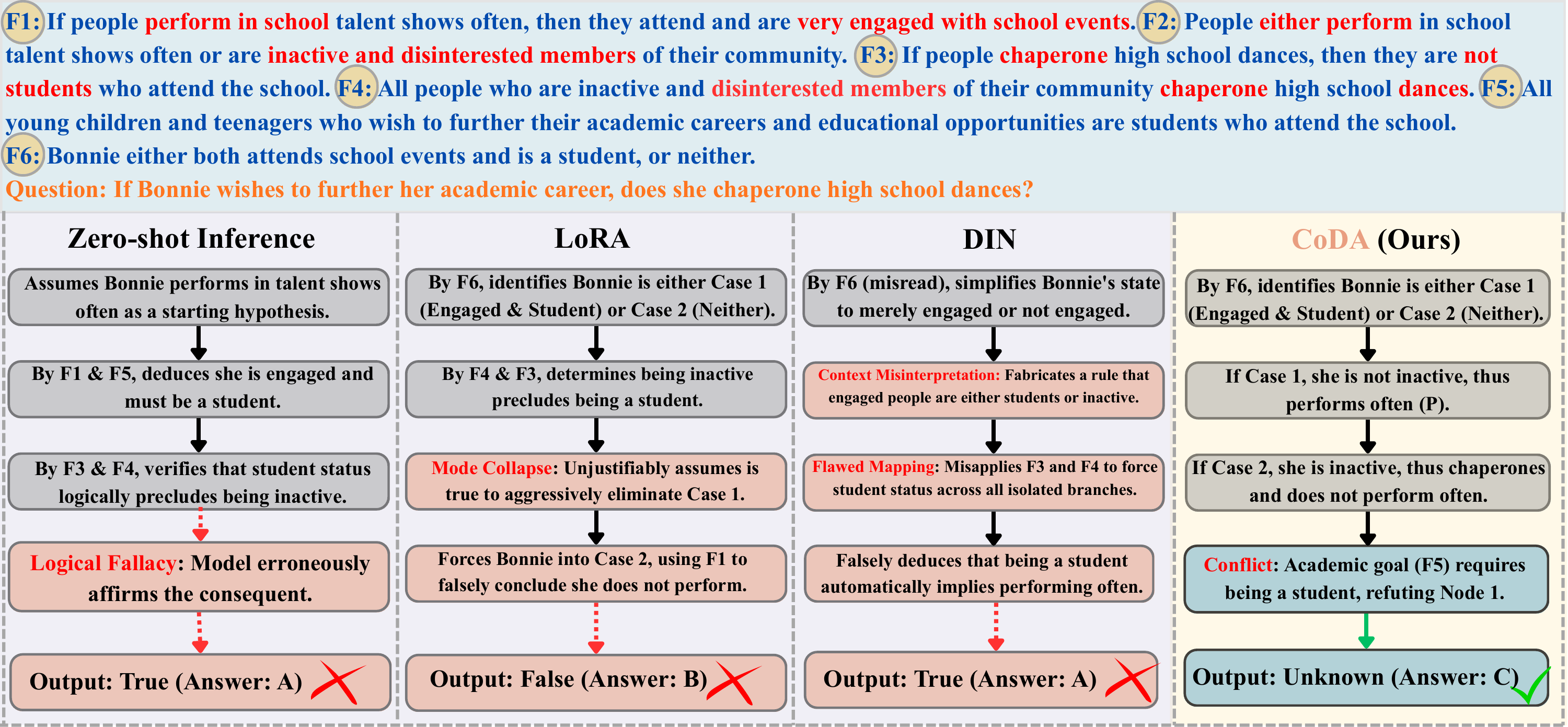}
  \label{case}
\vspace{-0.3in}
\end{figure*}

Furthermore, the effectiveness of our method consistently scales as the model size increases. For example, scaling from 12B to 27B within the Gemma-3-it family, our method's average performance rises from 81.3\% to 85.0\%. We observe a similar trend in the Qwen-2.5-Instruct series, where scaling from 14B to 32B increases the average score from 77.0\% to 80.7\%. We attribute this advantage to the fact that our proposed CoT-guided domain adapter can effectively identify and leverage the reasoning priors embedded in models of larger scales. We further conduct a case study to validate the efficacy of our method. As illustrated in Figure~\ref{case}, existing baselines struggle to robustly resolve complex multi-hop dependencies. Notably, zero-shot inference tends to produce unfounded inferential steps , whereas LoRA is severely hindered by mode collapse, which inappropriately disregards established logical premises. Furthermore, DIN is frequently misled by semantic noise from retrieved cross-domain exemplars, leading to erroneous structural mappings. In contrast, the \AdaSteer framework consistently maintains structural coherence through the robust transfer of underlying deductive reasoning patterns. By directly pinpointing the primary logical contradictions, such as the deductive conflict between Fact 3 and Fact 5, the method successfully bypasses superficial domain discrepancies to yield the correct inference.

\subsection{Domain Alignment Effectiveness (RQ2)}
\label{rq2}

\begin{figure}[t!]
    \centering
    \includegraphics[width=\linewidth]{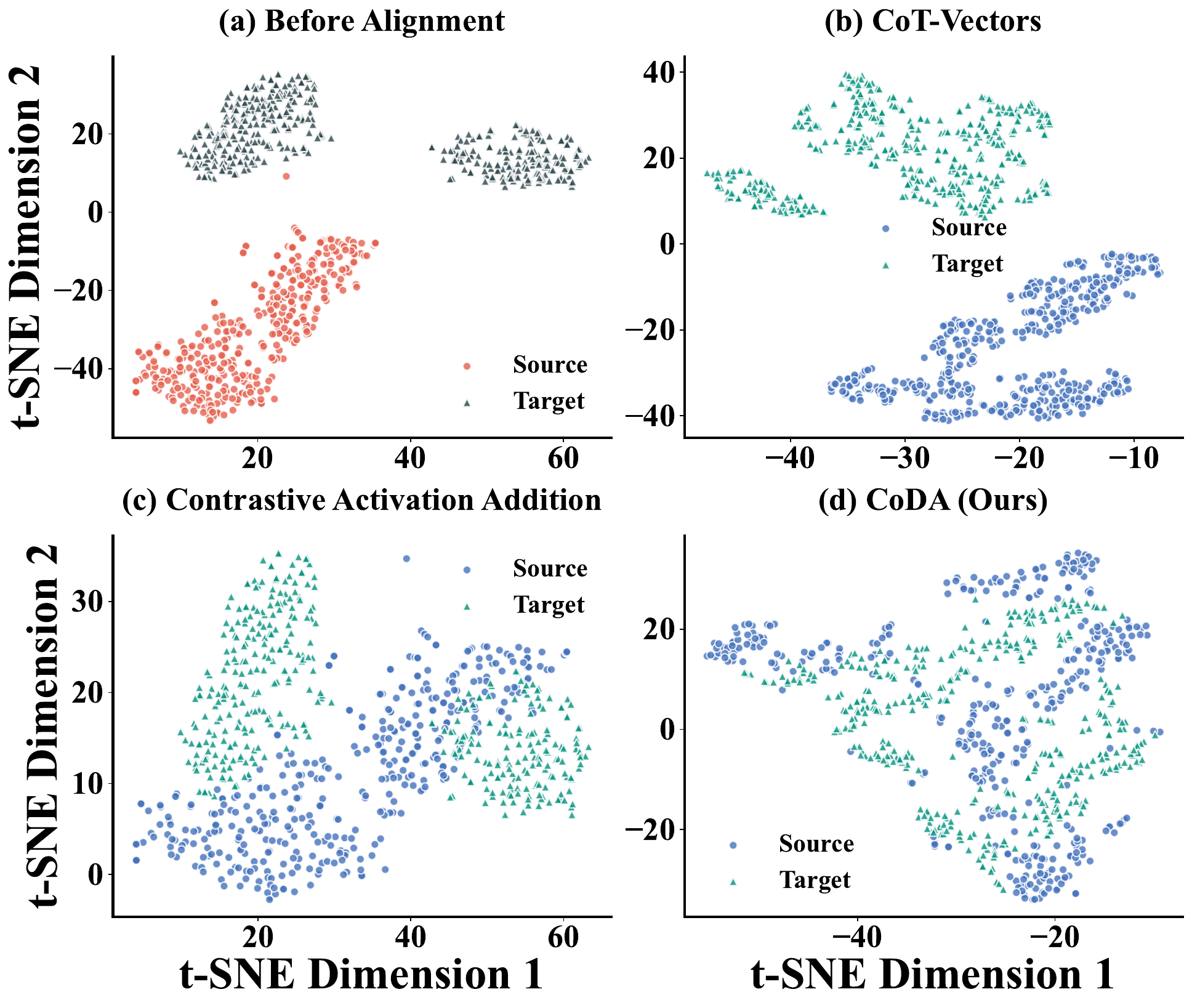}
    \caption{t-SNE visualization of latent representations. Compared to the unaligned state (a) and baselines (b, c) which exhibit domain separation or distortion, our \AdaSteer (d) achieves optimal feature fusion.}
    \label{fig:tsne}
\end{figure}

To validate the effectiveness of \AdaSteer stems from better alignment of source and target domains, we project the latent representations of the source and target domains into a two-dimensional space using t-SNE~\cite{van2008visualizing}. As shown in Figure~\ref{fig:tsne}, severe distribution shifts initially isolate the domains into distinct clusters, which is a primary cause of out-of-domain degradation in standard fine-tuning. \AdaSteer mitigates this isolation by aligning the distributions to establish a shared reasoning manifold for zero-shot transfer. Table~\ref{tab:tsne_metrics} quantifies this convergence. Unaligned states exhibit high isolation (Silhouette: 0.5998~\cite{rousseeuw1987silhouettes}, $k$-NN mixing: 0.12\%~\cite{buttner2019test}). CoT-Vectors reduce the global MMD (0.0358) but fail at local integration (0.00\% mixing), whereas Contrastive Activation Addition achieves 9.08\% mixing. \AdaSteer outperforms these baselines by jointly minimizing MMD (0.0354) and the Silhouette score (0.0321) while maximizing $k$-NN mixing (12.92\%), indicating robust manifold fusion at both the macro and instance levels.

\begin{table}[t!]
\centering
\small
\caption{Quantitative evaluation of latent space alignment on the 2D t-SNE projections. We report Silhouette Score ($\downarrow$) and 2D MMD ($\downarrow$) to measure macroscopic distribution distance, alongside k-NN Mixing Ratio ($\uparrow$, $k=10$) for microscopic instance-level fusion.}
\resizebox{0.4\textwidth}{!}{
\begin{tabular}{lccc}
\toprule
\textbf{Method} & \textbf{Silh.} $\downarrow$ & \textbf{Mix (\%)} $\uparrow$ & \textbf{MMD} $\downarrow$ \\
\midrule
Before Align  & 0.5998 & 0.12  & 0.0492 \\
CoT-Vectors   & 0.6527 & 0.00  & 0.0358 \\
CAA           & 0.1117 & 9.08  & 0.0391 \\
\AdaSteer & \bc \bf 0.0321 & \bc \bf 12.92 & \bc \bf 0.0354 \\
\bottomrule
\end{tabular}
}
\vspace{-0.2in}
\label{tab:tsne_metrics}
\end{table}

\subsection{Parameter Efficiency (RQ3)}

\begin{figure}[t!]
    \centering
    \includegraphics[width=\linewidth]{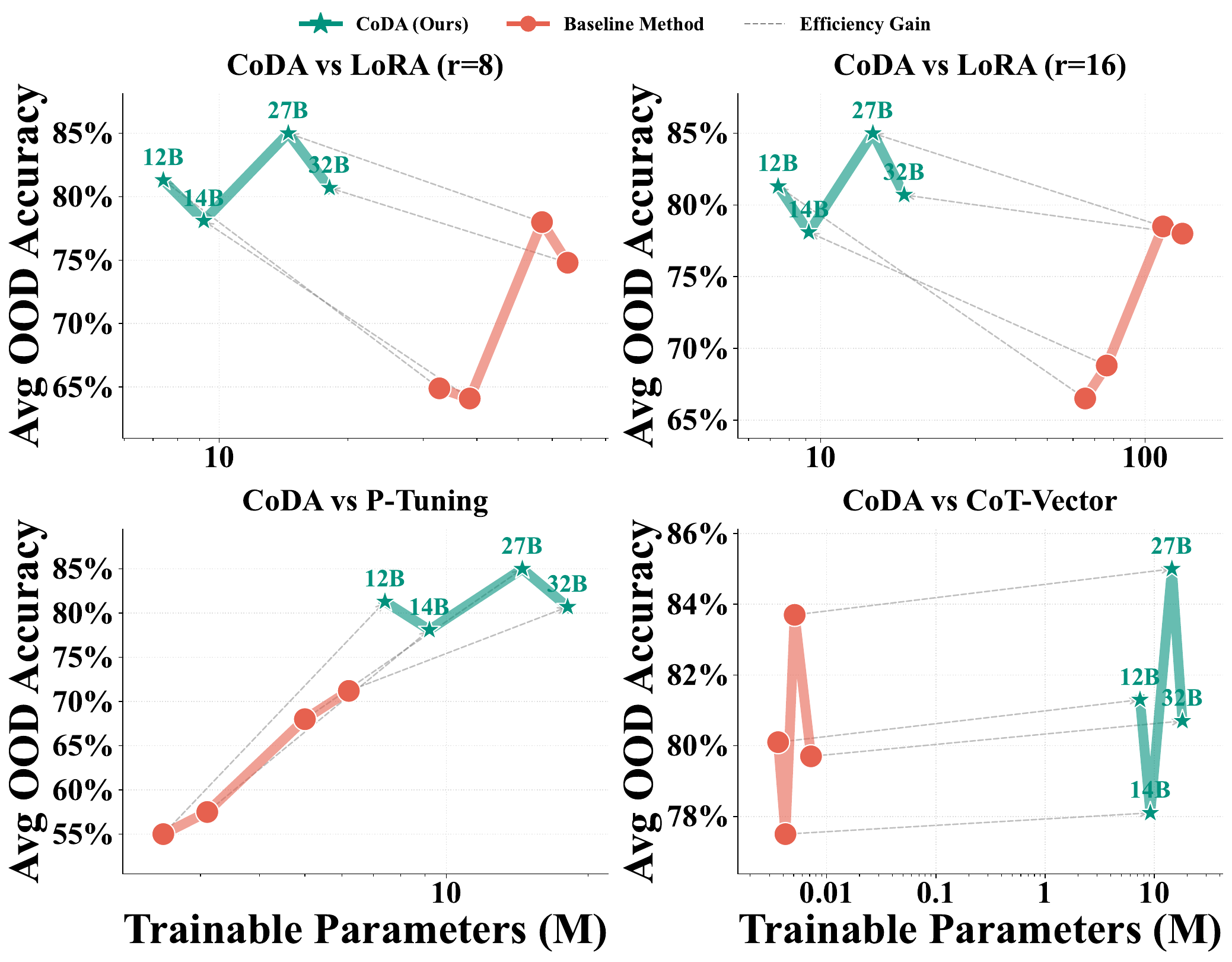}
    \caption{Parameter efficiency vs. OOD accuracy. Gray arrows highlight \AdaSteer's efficiency gains, consistently achieving higher accuracy with fewer trainable parameters than all baselines.}
    \label{fig:parameter_efficiency}
\end{figure}

We further analyze the trainable parameters to assess the efficiency of the method. As illustrated in Figure~\ref{fig:parameter_efficiency}, \AdaSteer improves the trade-off between parameter efficiency and out-of-domain generalization compared to standard LoRA across the evaluated 12B, 14B, 27B, and 32B model scales. At the 12B scale, LoRA uses 32.7M parameters but suffers from mode collapse, yielding only 64.0\% accuracy. \AdaSteer achieves 80.3\% accuracy with 7.4M parameters, improving the parameter-to-performance ratio from 1.96 to 10.85. This efficiency consistently scales to larger architectures. For instance, on the 27B model, \AdaSteer attains 82.9\% accuracy using 14.5M parameters, outperforming LoRA (76.0\% accuracy, 56.8M parameters). The results suggest that CoDA provides a more parameter-efficient and effective solution than conventional PEFT which even requires considerable amount of in-domain labeled data.

\subsection{Hyperparameter Analysis \& Ablation Study (RQ4)}
\label{rq3}

\begin{table*}[t!]
\centering
\small
\caption{Ablation study of \AdaSteer across different backbones (\texttt{Qwen2.5-32B-Instruct} and \texttt{Gemma3-27B-it}). We report the impact of removing the MMD penalty ($\mathcal{L}_{MMD}$) and the reasoning loss ($\mathcal{L}_{Reason}$) on cross-domain transfer performance.}
\resizebox{1.0\textwidth}{!}{
\begin{tabular}{llccccccccc}
\toprule
\textbf{Backbone Model} & \textbf{Ablation} & \textit{PW$\rightarrow$LD} & \textit{LD$\rightarrow$PW} & \textit{FO$\rightarrow$LD} & \textit{LD$\rightarrow$FO} & \textit{PW$\rightarrow$FO} & \textit{FO$\rightarrow$PW} & \textit{GS$\rightarrow$LD} & \textit{CS$\rightarrow$LD} & Avg. \\
\midrule
\multirow{3}{*}{\textbf{Qwen2.5-32B-Instruct}}
& \AdaSteer (Full) & \bc 88.0 & \bc 71.6 & \bc 88.3 & \bc 76.4 & \bc 75.9 & \bc 71.8 & \bc 87.0 & \bc 86.6 & \avgbg \bc \bf 80.7 \\
& - $\mathcal{L}_{MMD}$ & 84.5 & 68.2 & 85.1 & 72.8 & 72.5 & 68.5 & 83.0 & 82.5 & \avgbg 77.1 \\
& - $\mathcal{L}_{Reason}$ & 82.0 & 66.5 & 83.0 & 70.5 & 69.8 & 66.0 & 80.5 & 79.8 & \avgbg 74.8 \\
\midrule
\multirow{3}{*}{\textbf{Gemma3-27B-it}}
& \AdaSteer (Full) & \bc 91.6 & \bc 81.0 & \bc 87.0 & \bc 76.4 & \bc 79.1 & \bc 82.8 & \bc 91.2 & \bc 91.0 & \avgbg \bc \bf 85.0 \\
& - $\mathcal{L}_{MMD}$ & 88.2 & 77.5 & 83.5 & 73.0 & 75.8 & 79.5 & 88.0 & 87.5 & \avgbg 81.6 \\
& - $\mathcal{L}_{Reason}$ & 85.5 & 74.2 & 80.6 & 70.5 & 72.5 & 76.0 & 84.5 & 83.8 & \avgbg 78.4 \\
\midrule
\multirow{3}{*}{\textbf{Qwen2.5-14B-Instruct}}
& \AdaSteer (Full) & \bc 83.6 & \bc 66.1 & \bc 81.6 & \bc 76.4 & \bc 75.9 & \bc 67.1 & \bc 82.8 & \bc 82.5 & \avgbg \bc \bf 77.0 \\
& - $\mathcal{L}_{MMD}$ & 80.1 & 62.7 & 78.4 & 72.8 & 72.5 & 63.8 & 78.8 & 78.4 & \avgbg 73.4 \\
& - $\mathcal{L}_{Reason}$ & 77.6 & 61.0 & 76.3 & 70.5 & 69.8 & 61.3 & 76.3 & 75.7 & \avgbg 71.1 \\
\midrule
\multirow{3}{*}{\textbf{Gemma3-12B-it}}
& \AdaSteer (Full) & \bc 85.0 & \bc 78.5 & \bc 86.0 & \bc 74.5 & \bc 77.7 & \bc 80.5 & \bc 84.2 & \bc 84.3 & \avgbg \bc \bf 81.3 \\
& - $\mathcal{L}_{MMD}$ & 81.6 & 75.0 & 82.5 & 71.1 & 74.4 & 77.2 & 81.0 & 80.8 & \avgbg 77.9 \\
& - $\mathcal{L}_{Reason}$ & 78.9 & 71.7 & 79.6 & 68.6 & 71.1 & 73.7 & 77.5 & 77.1 & \avgbg 74.7 \\
\bottomrule
\end{tabular}
}
\vspace{-0.15in}
\label{tab:ablation}
\end{table*}

\begin{figure}[t!]
    \centering
    \includegraphics[width=\linewidth]{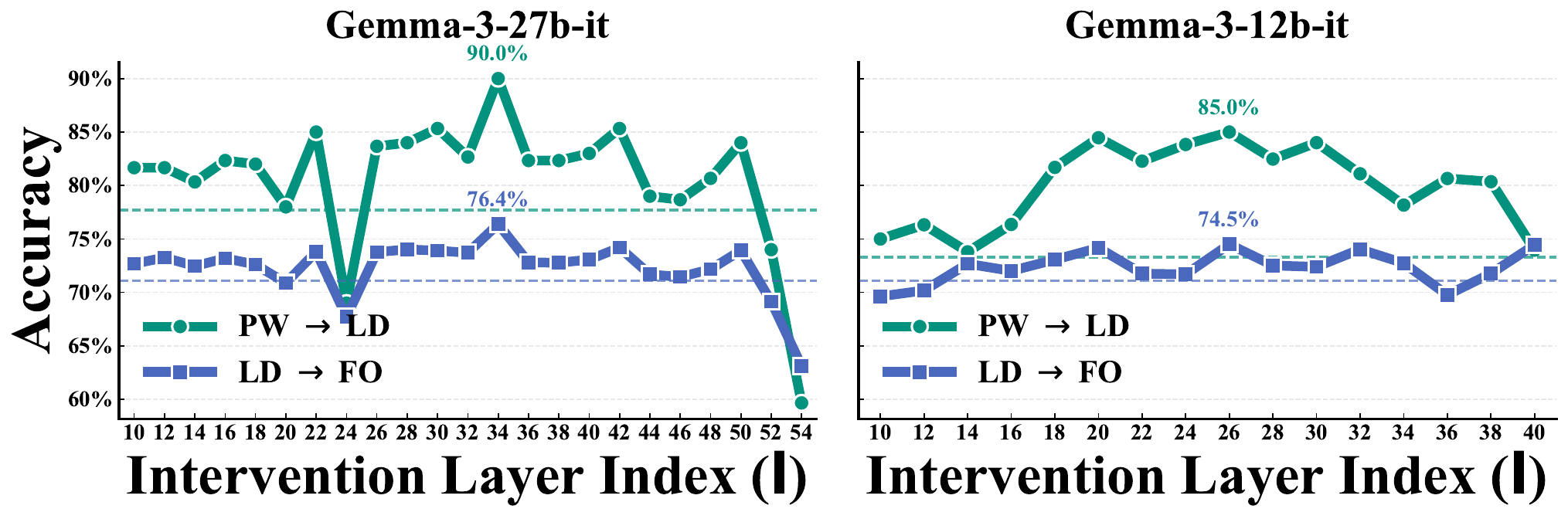}
    \caption{Layer-wise intervention performance. Accuracy of latent intervention at layer $l$ on Gemma-3-27b-it (left) and 12b-it (right). Dashed lines denote the zero-shot baselines for PW $\rightarrow$ LD (teal) and LD $\rightarrow$ FO (blue) tasks.}
    \label{fig:self_consistency}
\end{figure}

\begin{figure}[t]
    \centering
    \includegraphics[width=\linewidth]{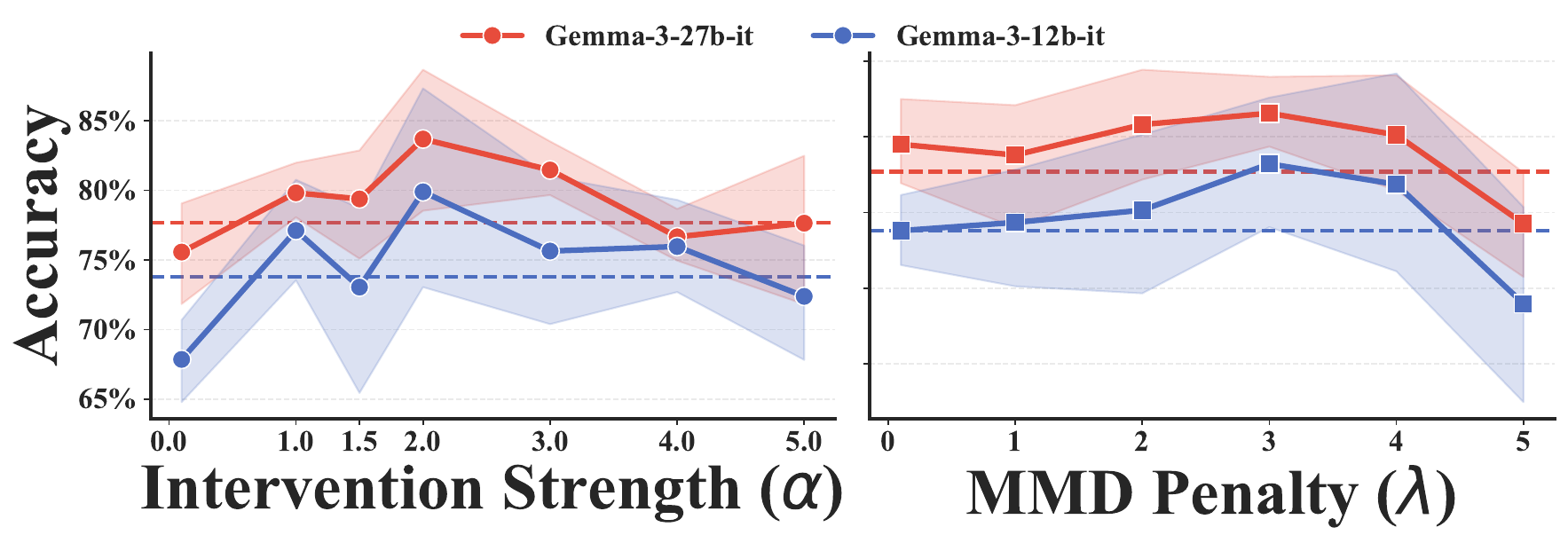}
    \caption{Hyperparameter sensitivity of Gemma-3 models. Accuracy variations across varying intervention strengths $\alpha$ (left) and MMD penalties $\lambda$ (right). Baselines are indicated by dashed lines.}
    \label{fig:sensitivity}
\end{figure}

For hyperparameter analysis, we first analyze the layer index $l$, which determines the optimal intervention space within the model architecture. The results are shown in Figure~\ref{fig:self_consistency}. Concretely, the effectiveness of latent intervention depends heavily on the layer index. Reasoning accuracy peaks at 90.0\% at layer 34, whereas interventions at shallow (layer 24) or terminal (layer 50+) stages lead to performance degradation. This pattern suggests that the optimal reasoning manifold resides in mid-to-late transformer blocks, as early layers encode shallow features and final layers tend to overfit the source domain distribution. We then analyze the intervention magnitude ($\alpha$) and the MMD alignment penalty ($\lambda$), which respectively determine the intensity of the injected reasoning signals during inference and the relative emphasis on cross-domain distribution matching during optimization. The results are shown in Figure~\ref{fig:sensitivity}. Concretely, a sensitivity analysis of these parameters reveals an inverted-U trend. Moderate strengths ($1.0 \le \alpha \le 2.5$) provide consistent improvements, but insufficient steering ($\alpha < 0.5$) fails to redirect the reasoning trajectory effectively.

To demonstrate the imperativeness of each module, we conduct a detailed ablation analysis. The results are shown in Table~\ref{tab:ablation}. Specifically, we can see that combining the reasoning loss ($\mathcal{L}_{Reason}$) and the MMD penalty ($\mathcal{L}_{MMD}$) is necessary for stable performance; removing either component reduces the model's ability to maintain a shared reasoning manifold.

\section{Related Work}

\subsection{Generalization in LLM}
Large language models (LLMs) often suffer substantial performance degradation under unforeseen domain shifts \cite{oncel2024adaptation, oh2025understanding, oh2024dawin, wu2025spine}. Existing approaches such as data-centric adaptation \cite{wang2024sharing, kargaran2025programming}, prompt calibration \cite{zhao2021calibrate, honda2025exploring, he2024using}, and parameter-efficient tuning \cite{hu2022lora} primarily modify external inputs or superficial structures, largely overlooking the inherent transferability encoded within their extensive pre-trained parameters. Consequently, recent work has begun systematically examining cross-domain representation alignment \cite{aghajanyan2021intrinsic, xu2026parameter}, including fine-grained neuron-level alignment in multilingual settings \cite{huang2025neurons, liu2025relation}. By directly manipulating the continuous latent space, these preliminary studies highlight the compelling potential of intrinsic model modulation. However, seamlessly applying these static structural insights to perform real-time, targeted interventions during dynamic inference remains a critical open challenge.

\subsection{Chain-of-Thought Distillation in LLMs}
Chain-of-Thought (CoT) distillation has emerged as a highly effective transfer learning paradigm for large language models, designed to instill complex problem-solving capabilities into compact student architectures \cite{magister2023teaching, ho2023large, hsieh2023distilling}. Building upon foundational knowledge distillation principles \cite{hinton2015distilling, gou2021knowledge}, this approach explicitly focuses on distilling the step-by-step reasoning trajectories generated by massive teacher models. Current methodologies typically achieve this transfer by fine-tuning the student model directly on teacher-generated rationales or by aligning the student's output distribution with the teacher's soft labels \cite{fu2023specializing, mukherjee2023orca}. While these techniques successfully condense deductive reasoning skills into smaller parameter spaces, conventional CoT distillation remains bottlenecked by computationally intensive parametric updates. Furthermore, it intrinsically assumes the availability of annotated or teacher-generated reasoning data within the target domain, a constraint that severely limits its applicability in zero-shot, cross-domain adaptation scenarios.

\subsection{LLM Reasoning}
Large Language Models (LLMs) have achieved substantial advancements across complex reasoning domains, largely driven by Chain-of-Thought (CoT) paradigms \cite{wei2022chain}. By decomposing intricate queries into sub-tasks, CoT empowers models to resolve multi-hop dependencies difficult to capture via direct input-to-output mapping \cite{huang2023towards}. Despite these breakthroughs, LLMs still lag behind human performance in systematic logical deduction tasks requiring rigorous step-by-step reasoning \cite{JIANG2026108924}. To bridge this gap, In-Context Learning (ICL) frequently augments inputs with expert-curated in-domain exemplars \cite{brown2020language, luo2024context}. Recent efforts extend this by retrieving structurally compatible cross-domain demonstrations as surrogates when specialized in-domain data is unavailable \cite{yan2026effectiveincontextcrossdomainknowledge, liu2026reasonanalogicallycrossdomainprior}. However, resulting performance gains remain modest due to pronounced domain shifts between source and target distributions \cite{tang2023large}. Under raw textual prompting, LLMs struggle to systematically abstract and transfer cross-domain knowledge \cite{siska2024examining}. This highlights the need for frameworks bypassing surface-level discrepancies to align disparate representation spaces through deeper latent interventions.

\section{Conclusion}

In this work, we introduce \AdaSteer, a parameter-efficient framework for cross-domain knowledge transfer of large language models. By treating domain adaptation as a latent space intervention, \AdaSteer uses a lightweight domain adapter to embed cross-domain knowledge and intervene in the
intermediate hidden states. Jointly optimizing a reasoning loss and a maximum mean discrepancy (MMD) alignment penalty bridges the representational gap between source and target domains without requiring target-domain labels. Extensive experimental results across multiple LLM families show that \AdaSteer avoids the mode collapse which commonly exists in standard fine-tuning approaches like LoRA, improving reasoning accuracy while maintaining parameter efficiency. This approach offers a practical method for transferring the reasoning capabilities of LLMs to unannotated target distributions. Future work will leverage mechanistic interpretability (e.g., Sparse AutoEncoder and attention attribution) to explicitly decode the logical features manipulated by these latent interventions.

\newpage
\bibliographystyle{IEEEtran}
\bibliography{reference}

\vfill

\end{document}